\documentclass[letterpaper, 10 pt, conference]{ieeeconf}  % Comment this line out if you need a4paper

\IEEEoverridecommandlockouts                              % This command is only needed if 
\usepackage{graphicx}
\usepackage{amsfonts}
\usepackage{amsmath}
\usepackage{amssymb}
\usepackage{bm}
\usepackage{marvosym}
\usepackage{indentfirst}
\usepackage{multirow, makecell}
\usepackage{booktabs}
\usepackage{tabu}
\usepackage[normalem]{ulem}
\usepackage[table,xcdraw]{xcolor}
\usepackage[section]{placeins}
\usepackage{rotfloat}
\usepackage{graphicx}
\makeatletter
\let\NAT@parse\undefined
\makeatother
\usepackage[colorlinks,linkcolor=blue,citecolor=green]{hyperref}

\title{\LARGE \bf
UnPWC-SVDLO: Multi-SVD on PointPWC for Unsupervised Lidar Odometry
}

\author{Yiming Tu}

\begin{document}

\maketitle
\thispagestyle{empty}
\pagestyle{empty}

%%%%%%%%%%%%%%%%%%%%%%%%%%%%%%%%%%%%%%%%%%%%%%%%%%%%%%%%%%%%%%%%%%%%%%%%%%%%%%%%
\begin{abstract}
High-precision lidar odomety is an essential part of autonomous driving.
In recent years, deep learning methods have been widely used in lidar odomety 
tasks, but most of the current methods only extract the global features of the 
point clouds. It is impossible to obtain more detailed point-level features in 
this way. In addition, only the fully connected layer is used to estimate 
the pose. The fully connected layer has achieved obvious results in the 
classification task, but the changes in pose are a continuous rather than 
discrete process, high-precision pose estimation can not be obtained only 
by using the fully connected layer. Our method avoids problems mentioned above. 
We use PointPWC \cite{PointPWC} as our backbone network. PointPWC \cite{PointPWC} 
is originally used for scene flow estimation. The scene flow estimation task has 
a strong correlation with lidar odomety. Traget point clouds can be 
obtained by adding the scene flow and source point clouds. We can achieve the 
pose directly through ICP algorithm \cite{po2plICP} solved by SVD, and the fully 
connected layer is no longer used. PointPWC \cite{PointPWC} extracts 
point-level features from point clouds with different sampling levels, which 
solves the problem of too rough feature extraction. We conduct experiments on
KITTI \cite{KITTI},  Ford Campus Vision and Lidar DataSet \cite{Ford} and 
Apollo-SouthBay Dataset \cite{Apollo}. Our result is comparable with the 
state-of-the-art unsupervised deep learing method SelfVoxeLO \cite{SelfVoxeLO}.

\end{abstract}

%%%%%%%%%%%%%%%%%%%%%%%%%%%%%%%%%%%%%%%%%%%%%%%%%%%%%%%%%%%%%%%%%%%%%%%%%%%%%%%%

\section{INTRODUCTION}
\begin{figure}[!t]
        \includegraphics[width=\linewidth]{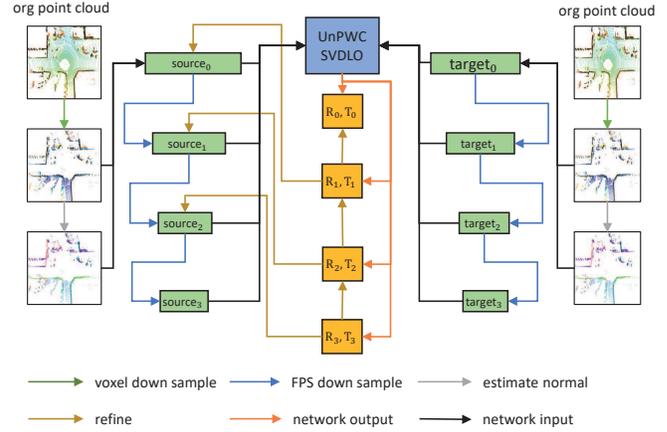}
        \caption{The overall structure of our framework, the original point clouds 
        first voxel down sampling and calculate normal vector of downsampled point 
        clouds, then we obtain multi-level point clouds by multi-level downsample. 
        The high-level point clouds first input the network to calculate the pose, 
        and then the low-level point clouds is refined. The low-level poses are 
        merged with network output poses to obtain the high-level poses.} 
        \label{fig1}
    \end{figure}

High-precision pose estimation is the core of SLAM, and now it occupies an 
important position in the field of robotics and autonomous driving. Because 
traditional methods have relatively low requirements for computer hardware, they 
are the first to be applied to lidar odometry. As the most classic point clouds 
registration algorithm, ICP \cite{po2plICP} is directly applied to lidar odometry 
tasks, but its performance is not satisfactory. After that, ICP-based algorithm LOAM 
\cite{Loam} performs a variety of preprocessing on the original point clouds, and 
then uses Gauss-Newton iteration method to obtain the pose after extracting the key points, 
which makes the accuracy of pose estimation a big step forward. With the development 
of computer hardware, deep learning methods that were once slow can be applied to 
odomety tasks that require real-time performance. Due to the irregularity of the 
point clouds, most of the deep lidar odometry methods project point clouds to the 2D 
plane \cite{Lo-net,Velas,Cho,Nubert,DeepPCO} or 3D voxel \cite{SelfVoxeLO} 
according to the coordinates, and use the 2/3D convolutional 
neural network to extracts global features by normalized data. \\
\indent In order to avoid the roughness of global features, our method considers 
extracting point-level features to estimate pose. Inspired by the high similarity 
of secen flow estimation and lidar odometry, we use PointPWC \cite{PointPWC} as 
the backbone network to extract point-level features and convert sence flow into poses 
through SVD (see Figure.~\ref{fig1}). PointPWC \cite{PointPWC} adopts the L-level pyramid of point 
structure, and each level of point clouds estimated secen flow, so 
L-level poses can be generated. At the same time, the high-level poses can refine 
the low-level point clouds. Except for the highest-level pose, the rest poses 
obtained at each level is the residual pose. The final pose of each level is 
obtained by fusing the residual pose and the final pose of the higher level. 
Finally, the point-to-plane ICP algorithm is used as the loss function to correct 
the pose of each level. The entire training process is unsupervised. We compared 
many existing traditional and deep lidar odometry methods on KITTI \cite{KITTI},  
Ford Campus Vision and Lidar Data Set \cite{Ford} and Apollo-SouthBay Dataset 
\cite{Apollo}, and achieved good results. Our main contributions are as follows: \\
\begin{itemize}
\item We abandon deep learning framework that directly extracts global 
features from the entire point clouds to fit the pose. Instead, we choose to extract 
point-level features and output scene flow for each level point. At last, we use SVD 
decomposition to obtain pose. \\
\item We use multi-layer pose estimation. High-level pose is used to refine the 
lower-level points which make higher level pose has higher accuracy. \\
\item We train our network with unsupervised loss and surpass the best existing 
unsupervised methods on KITTI \cite{KITTI} and Apollo \cite{Apollo}, and can even 
be comparable  to the supervised method Lo-Net \cite{Lo-net}, and are only slightly 
lower than Lo-Net \cite{Lo-net} on Ford \cite{Ford}. 
\end{itemize}

\section{RELATED WORK}
\subsection{Deep LiDAR Odometry}
With the development of computer hardware, deep learning has continued to reduce 
its operating time, which is sufficient to meet the real-time requirements of 
lidar odometry. These deep learning methods can be divided into two categories: 
supervised and unsupervised. \\
\indent Supervised methods appear relatively early, Velas \emph{et al.} \cite{Velas} 
first map the point clouds s to 2D "image" by spherical projection and regard pose 
estimation as a classification problem. Lo-net \cite{Lo-net} also use spherical 
projection and add normal vector which calculated from the surrounding point clouds  
as network input. Wang \emph{et al.} \cite{DeepPCO} adopt a dual-branch 
architecture to infer 3-D translation and orientation separately instead of a 
single network. Differently, Li \emph{et al.} \cite{DMLO} use neural network to 
calculate the match between points and obtain pose by SVD. Use Pointnet 
\cite{PointNet} as backbone, Zheng \emph{et al.} \cite{LodoNet} propose a  
framework for extracting feature from matching keypoint pairs (MKPs) which 
extracted effectively and efficiently by projecting 3D point clouds s into 2D 
spherical depth images. PWCLO-Net \cite{PWCLO} first introduces the advantages 
of the scene flow estimation network into the lidar odometry, adds the feature extraction 
modules in PointPwc \cite{PointPWC} to its own network, and introduces the 
hierarchical embedding mask to remove the points that are not highly matched. \\ 
\indent Unsupervised methods appear later. Cho \emph{et al.} \cite{Cho} first 
apply unsupervised approach on deep-learning-based lidar odometry which is 
an extension of their previous approach \cite{DeepLo}. The inspiration of its 
loss function comes from point-to-plane ICP \cite{po2plICP}. Then, Nubert 
\emph{et al.} \cite{Nubert} report methods with similarly models, but they use 
different way to calculate normals and find matching points. In addition, they
add plane-to-plane ICP loss. SelfVoxeLO \cite{SelfVoxeLO} use 3D voxel to 
downsample and normalize the point clouds, and use 3D convolution network 
to extract the features of voxel. they also put forward many innovative loss 
functions to greatly improve the accuracy of pose.

\subsection{Scene Flow Estimation}
Scene flow estimation and lidar odometry are closely related. They also need to 
find the matching relationship between a pair of points in the point clouds. The 
purpose of scene flow estimation is to accurately find the changing trend of each 
point, while lidar odometry is to get the entire point. The change trend between 
point clouds, but its change trend has only 3-DoF, which is much 
easier than the change of the 6-DoF of the lidar odometry. These 
two tasks have their own difficulties, but they can transform each other. \\
\indent FlowNet3D \cite{Flownet3d} based on PointNet++, use embeding layers to 
learn the motion between points in two consecutive frames. PointPWC \cite{PointPWC} 
propose the cost volume method on point clouds and use a muti-level pyramid 
structure network to estimate muti-level scene flow. Also they first
introduce self-supervised losses in sence flow estimate task.

\section{PROPOSED APPROACH}
\subsection{Problem discription and Data Process}
For each timestamp $k\in\mathbb{R}^+$, we will get a frame of point cloud $P_k$ 
as input, and the output will be the pose transformation between point clouds 
$P_k$ and $P_{k+1}$. Since we use PointPWC \cite{PointPWC} as the backbone 
network, the KNN points of each point in point clouds are calculated multiple 
times, if the number of point clouds is too large, We need to spend a lot of 
time to get results , which cannot meet the real-time requirements of odometry.
\label{sec:down} The point cloud is sorted according to the z-axis first
, and the $U\%$ points with the smaller z-axis are removed (roughly remove the 
ground points, the ground points will also be removed in the preprocessing step 
in the scene flow estimation task). then we perform voxel downsample on point clouds 
by dividing the space into equal-sized cells whose side length is $D$ which follows 
SelfVoxeLO \cite{SelfVoxeLO}. For each cell, the arithmetic average of all points in 
it is used to represent it. We name downsampled point cloud
as $S_k^0\in\mathbb{R}^{C_0\times3}$. At last, the normal vector 
$N_k^0\in\mathbb{R}^{C_0\times3}$ of $S_k^0$ is calculated by the 
plane fitting method \cite{norestimate}.  
\begin{figure}[!t]
        \includegraphics[width=\linewidth]{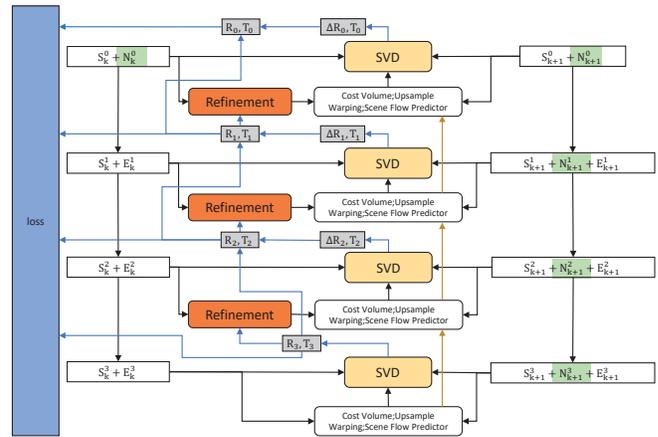}
        \caption{The various modules of our network. Colored modules and data 
        are part of our innovation, and others are originally part 
        in PointPWC \cite{PointPWC}. There are 4 levels in total. Level-0 
        represents the original point cloud and the final output pose. Level-3 
        is the highest sampled point cloud, and the pose is directly output 
        through SVD without fusion with the higher level.} 
        \label{fig2}
\end{figure}
\begin{figure}[!t]
        \includegraphics[width=\linewidth]{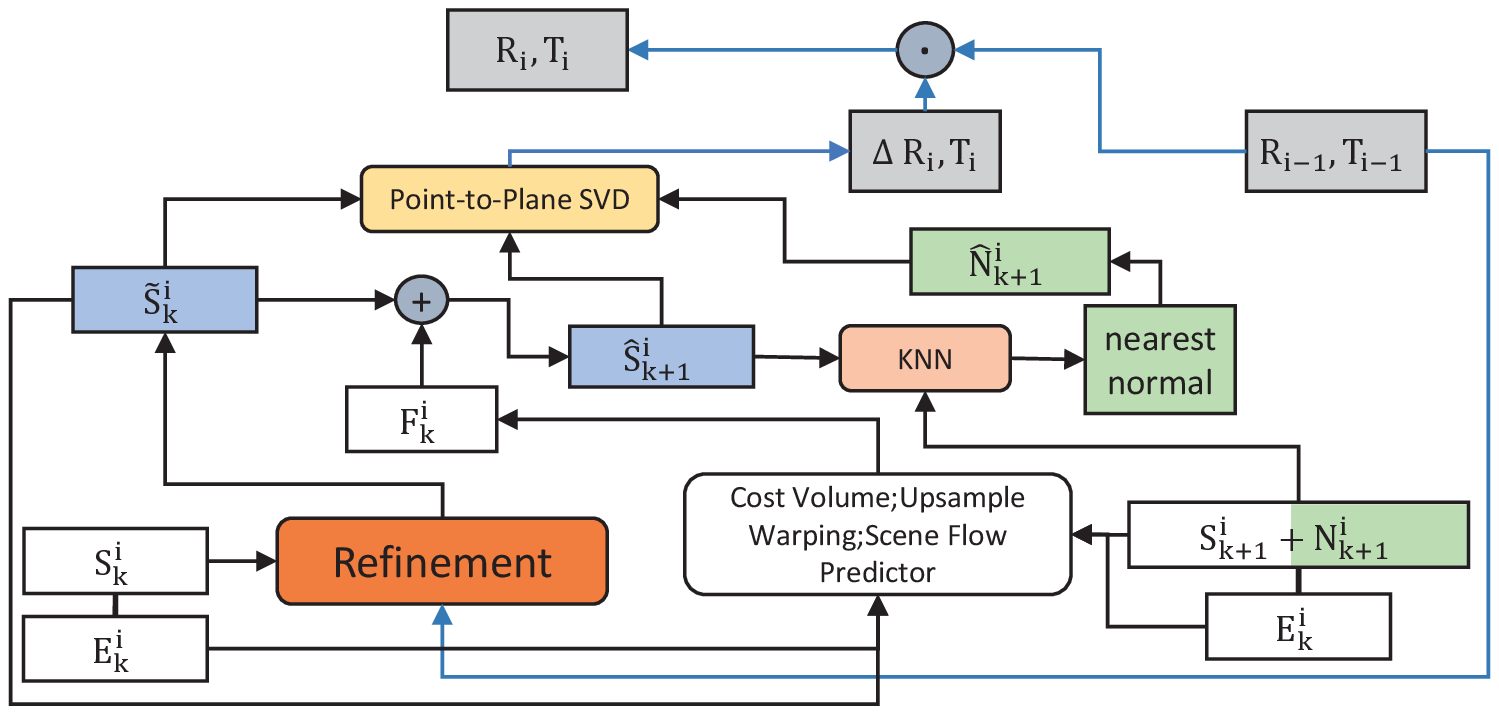}
        \caption{Level-i data processing details.} 
        \label{fig3}
\end{figure}
\begin{figure}[!t]
        \includegraphics[width=\linewidth]{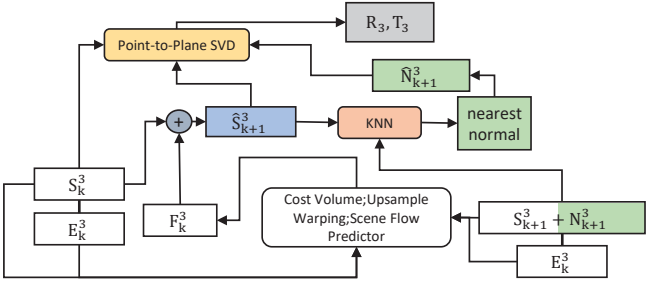}
        \caption{Level-3 data processing details.} 
        \label{fig4}
\end{figure}
\subsection{Network Architecture}
The overall structure  of the network is shown in 
Figure.~\ref{fig2}. We first perform multi-level FPS (Farthest Point Sample) and feature 
extraction on the original point cloud $S_k^0$, and then start up-sampling 
and fuse features from the highest Level-3 $S_k^3$ to estimate the pose of each level. 
\subsubsection{FPS and Feature Extract}
We input $S_k^0$ and $N_k^0$ into the network, and perform multi-level 
FPS on $S_k^0$. $S_k^{i+1}$ is FPS from 
$S_k^i\in\mathbb{R}^{C_i\times3}$. For each $S_k^i$, we can get corresponding 
$N_k^i\in\mathbb{R}^{C_i\times3}$. While down-sampling, PointConv 
\cite{Pointconv} is used to extract the feature $E_k^i\in\mathbb{R}^{C_i\times M_i}$ 
of $S_k^i$ which is represented as
\begin{equation}                  
E_k^i=
\begin{cases}              
        PointConv(N_k^{i-1}) & i = 1\\  
        PointConv(E_k^{i-1}) & i > 1\\  
\end{cases},     
\end{equation}
so that we can get the multi-level data $(S_k^i,N_k^i,E_k^i)$. Then start from 
the highest sampling Level-3 ($S_k^3$ with the least number of points in 
$S_k^i$). we perform up-sampling and solve the pose. 
\subsubsection{Up-sampling, Feature Fusion and SVD Sovle the Pose}
The detail of this part is shown in Figure.~\ref{fig3} and Figure.~\ref{fig4}.
Except for the highest level of $S_k^3$, each level of source point cloud 
$S_k^i$ will first be refined by $R_{i+1}\in\mathbb{R}^{3\times 3}, 
T_{i+1}\in\mathbb{R}^{3\times 1}$(we show how to get them later) to obtain 
$\widetilde{S}_k^i$ by following formula
\begin{equation}
\widetilde{S}_k^i=
\begin{cases}              
        S_k^i & i = 3\\  
        R_{i+1}\cdot S_k^i+T_{i+1} & i < 3\\  
\end{cases}.
\end{equation}
Then we enter source data $(S_k^i,E_k^i)$ and target data 
$(S_{k+1}^i,E_{k+1}^i)$ into feature fusion part of PointPWC \cite{PointPWC} 
(Cost Volume; Upsample; Warping; Scene Flow Predictor) to obtain 
the scene flow $F_k^i$.  
$\widetilde{S}_k^i$ and $F_k^i\in\mathbb{R}^{C_i\times3}$ are added as the 
generated target point cloud $\widehat{S}_{k+1}^i$,
\begin{equation}
\widehat{S}_{k+1}^i=\widetilde{S}_{k}^i+F_k^i.
\end{equation}
By refined source point cloud $\widetilde{S}_k^i$, 
generated target point cloud $\widehat{S}_{k+1}^i$, target normal vector $N_{k+1}^i$, 
% scene flow $F_k^i$, 
we can obtain the 
pose $\Delta{R_i},\Delta{T_i}$ between $\widetilde{S}_k^i$ and $\widehat{S}_{k+1}^i$.
Each point in $\widehat{S}_{k+1}^i$ finds the nearest neighbor point in 
$S_{k+1}^i$ as the normal vector to obtain $\widehat{N}_{k+1}^i$, 
\begin{equation}
\begin{split}        
\widehat{N}_{k+1}^i=\{n\in N_{k+1}^i | \mathop{\min}\limits_
{s}(\widehat{s}-s)^2,\\
\widehat{s}\in \widehat{S}_{k+1}^i,
s\in S_{k+1}^i\}.
\end{split}
\end{equation}
According to \cite{po2plICP}, if euler angle of $R$ is closed to zero, 
we can simplify the formula of point-to-plane ICP and transform it 
into a standard linear least-squares problem as the following formula
\begin{equation}
\begin{split}         
\mathop{\min}\limits_{\Delta{R_i},\Delta{T_i}}&\sum_{j=1}^K((\Delta{R_i}\cdot\widetilde{s}_j
+\Delta{T_i}-\widehat{s}_j)\cdot\widehat{n}_j) \\         
&=\mathop{\min}\limits_{x}|A\cdot x-b|^2 \\
&\widehat{s}_j\in \widehat{S}_{k+1}^i,
\widehat{n}_j\in \widehat{N}_{k+1}^i,
\widetilde{s}_j\in \widetilde{S}_{k}^i, \\
& A\in \mathbb{R}^{C_i\times 6}, b\in \mathbb{R}^{C_i\times 1}, x\in \mathbb{R}^{6\times 1}
\end{split}
\end{equation}
where
\begin{equation}
b_j=\sum_c^{x,y,z}\widehat{n}_j^c\widehat{s}_j^c-\widehat{n}_j^c\widetilde{s}_j^c,     
\end{equation}
\begin{equation}
x=[r^x, r^y, r^z, t^x, t^y, t^z]^T     
\end{equation}
and
\begin{equation}
\begin{split}       
A_j=&[a_j^x, a_j^y, a_j^z, \widehat{n}_j^x, \widehat{n}_j^y, \widehat{n}_j^z] \\                
&a_j^x=\widehat{n}_j^z\widetilde{s}_j^y-\widehat{n}_j^y\widetilde{s}_j^z \\
&a_j^y=\widehat{n}_j^x\widetilde{s}_j^z-\widehat{n}_j^z\widetilde{s}_j^x \\
&a_j^z=\widehat{n}_j^y\widetilde{s}_j^x-\widehat{n}_j^x\widetilde{s}_j^y \\
\end{split}       
\end{equation}
which can be solved by using SVD as formula $A=U\Sigma V^T$. Next we can get the
pseudo-inverse of $A$ as $A^+=V\Sigma^+ U^T$. So best result of $x$ can be
achieved by
\begin{equation}
x_{opt}=A^+b.        
\end{equation}  
At last, We transform $x$ into $\Delta{R_i},\Delta{T_i}$ by
\begin{equation}
\Delta{T_i}=[t^x, t^y, t^z]        
\end{equation}
and
\begin{equation}
\begin{split}        
R^x=&\left[
\begin{array}{ccc}   
        1 & 0 & 0 \\
        0 & \cos r^x & -\sin r^x \\
        0 & \sin r^x & \cos r^x    
\end{array}\right]\\
R^y=&\left[
\begin{array}{ccc}   
        \cos r^y & 0 & \sin r^y \\
        0 & 1 &0  \\
        -\sin r^y & 0 & \cos r^y    
\end{array}\right]\\
R^z=&\left[
\begin{array}{ccc}   
        \cos r^z & -\sin r^z & 0 \\
        \sin r^z &  \cos r^z & 0 \\
        0 &  0&  1  
\end{array}\right]\\              
&\Delta{R_i}=R^z\cdot R^y\cdot R^x.
\end{split}
\end{equation}
Also we can obtain Level-i final pose $R_i,T_i$ by
\begin{equation}
R_i=
\begin{cases}              
        \Delta{R_i} & i = 3\\  
        \Delta{R_i}\cdot R_{i+1} & i < 3\\  
\end{cases}       
\end{equation}
and
\begin{equation}
T_i=
\begin{cases}              
\Delta{T_i} & i = 3\\  
\Delta{R_i}\cdot T_{i+1}+\Delta{T_i} & i < 3\\  
\end{cases} 
\end{equation}
\begin{table*}[!t]
\renewcommand\arraystretch{1.1}
\normalsize      
\setlength{\belowcaptionskip}{-0.6cm}
\setlength{\abovecaptionskip}{-0.5cm}
\caption{KITTI odometry evaluation.}\label{tab1}
\begin{center}
\resizebox{1.0\textwidth}{!}{       
\begin{tabular}{c|cc|cc|cc|cc|cc|cc|cc|cc|cc|cc|cc|cc}
\hline
                            & \multicolumn{2}{c|}{00*}                                                    & \multicolumn{2}{c|}{01*}                                  & \multicolumn{2}{c|}{02*}                                           & \multicolumn{2}{c|}{03*}                                  & \multicolumn{2}{c|}{04*}                                  & \multicolumn{2}{c|}{05*}                                                    & \multicolumn{2}{c|}{06*}                                                    & \multicolumn{2}{c|}{07}                                                     & \multicolumn{2}{c|}{08}                                                     & \multicolumn{2}{c|}{09}                                   & \multicolumn{2}{c|}{10}                                            & \multicolumn{2}{c}{Mean on 07-10}                                          \\ \cline{2-25} 
\multirow{-2}{*}{Method}    & $t_{rel}$                            & $r_{rel}$                            & $t_{rel}$                   & $r_{rel}$                   & $t_{rel}$                   & $r_{rel}$                            & $t_{rel}$                   & $r_{rel}$                   & $t_{rel}$                   & $r_{rel}$                   & $t_{rel}$                            & $r_{rel}$                            & $t_{rel}$                            & $r_{rel}$                            & $t_{rel}$                            & $r_{rel}$                            & $t_{rel}$                            & $r_{rel}$                            & $t_{rel}$                   & $r_{rel}$                   & $t_{rel}$                            & $r_{rel}$                   & $t_{rel}$                             & $r_{rel}$                    \\ \hline
ICP-po2po                   & 6.88                                 & 2.99                                 & 11.21                       & 2.58                        & 8.21                        & 3.39                                 & 11.07                       & 5.05                        & 6.64                        & 4.02                        & 3.97                                 & 1.93                                 & 1.95                                 & 1.59                                 & 5.17                                 & 3.35                                 & 10.04                                & 4.93                                 & 6.93                        & 2.89                        & 8.91                                 & 4.74                        & 7.763                                 & 3.978                        \\
ICP-po2pl                   & 3.8                                  & 1.73                                 & 13.53                       & 2.58                        & 9                           & 2.74                                 & 2.72                        & 1.63                        & 2.96                        & 2.58                        & 2.29                                 & 1.08                                 & 1.77                                 & 1                                    & 1.55                                 & 1.42                                 & 4.42                                 & 2.14                                 & 3.95                        & 1.71                        & 6.13                                 & 2.6                         & 4.013                                 & 1.968                        \\
GICP \cite{GICP}                       & 1.29                                 & 0.64                                 & 4.39                        & 0.91                        & 2.53                        & 0.77                                 & 1.68                        & 1.08                        & 3.76                        & 1.07                        & 1.02                                 & 0.54                                 & 0.92                                 & 0.46                                 & 0.64                                 & \textbf{0.45}                        & 1.58                                 & 0.75                                 & 1.97                        & 0.77                        & \textbf{1.31}                        & 0.62                        & 1.375                                 & 0.648                        \\
CLS \cite{CLS}                        & 2.11                                 & 0.95                                 & 4.22                        & 1.05                        & 2.29                        & 0.86                                 & 1.63                        & 1.09                        & 1.59                        & 0.71                        & 1.98                                 & 0.92                                 & 0.92                                 & 0.46                                 & 1.04                                 & 0.73                                 & 2.14                                 & 1.05                                 & 1.95                        & 0.92                        & 3.46                                 & 1.28                        & 2.148                                 & 0.995                        \\
LOAM(w/o mapping) \cite{Loam}          & 15.99                                & 6.25                                 & 3.43                        & 0.93                        & 9.4                         & 3.68                                 & 18.18                       & 9.91                        & 9.59                        & 4.57                        & 9.16                                 & 4.1                                  & 8.91                                 & 4.63                                 & 10.87                                & 6.76                                 & 12.72                                & 5.77                                 & 8.1                         & 4.3                         & 12.67                                & 8.79                        & 11.090                                & 6.405                        \\
LOAM(w/ mapping) \cite{Loam}           & \textbf{1.1}                         & \textbf{0.53}                        & \textbf{2.79}               & \textbf{0.55}               & \textbf{1.54}               & \textbf{0.55}                        & \textbf{1.13}               & \textbf{0.65}               & \textbf{1.45}               & \textbf{0.5}                & \textbf{0.75}                        & \textbf{0.38}                        & \textbf{0.72}                        & \textbf{0.39}                        & \textbf{0.69}                        & 0.5                                  & \textbf{1.18}                        & \textbf{0.44}                        & \textbf{1.2}                & \textbf{0.48}               & 1.51                                 & \textbf{0.57}               & \textbf{1.145}                        & \textbf{0.498}               \\ \hline
LO-Net \cite{Lo-net}                     & 1.47                                 & \textbf{0.72}                                 & \textbf{1.36}               & \textbf{0.47}               & 1.52               & 0.71                                 & \textbf{1.03}               & \textbf{0.66}               & \textbf{0.51}               & \textbf{0.65}               & \textbf{1.04}                                 & 0.69                                 & 0.71                                 & 0.5                                  & 1.7                                  & 0.89                                 & 2.12                                 & 0.77                                 & 1.37               & \textbf{0.58}               & \textbf{1.8}                                  & 0.93              & 1.748                                 & 0.793               \\
SelfVoxeLO \cite{SelfVoxeLO}                & NA                                   & NA                                   & NA                          & NA                          & NA                          & NA                                   & NA                          & NA                          & NA                          & NA                          & NA                                   & NA                                   & NA                                   & NA                                   & 3.09                                 & 1.81                                 & 3.16                                 & 1.14                                 & 3.01                        & 1.14                        & 3.48                                 & 1.11                        & 3.185                                 & 1.300                        \\
Nubert et al. \cite{Nubert}              & NA                                   & NA                                   & NA                          & NA                          & NA                          & NA                                   & NA                          & NA                          & NA                          & NA                          & NA                                   & NA                                   & NA                                   & NA                                   & NA                                   & NA                                   & NA                                   & NA                                   & 6.05                        & 2.15                        & 6.44                                 & 3.00                        & 6.245                                 & 2.575                        \\
Cho et al. \cite{Cho}                 & NA                                   & NA                                   & NA                          & NA                          & NA                          & NA                                   & NA                          & NA                          & NA                          & NA                          & NA                                   & NA                                   & NA                                   & NA                                   & NA                                   & NA                                   & NA                                   & NA                                   & 4.87                        & 1.95                        & 5.02                                 & 1.83                        & 4.945                                 & 1.890                        \\
{\color[HTML]{4472C4} Ours} & {\color[HTML]{4472C4} \textbf{1.36}} & {\color[HTML]{4472C4} 0.73} & {\color[HTML]{4472C4} 3.39} & {\color[HTML]{4472C4} 0.95} & {\color[HTML]{4472C4} \textbf{1.45}} & {\color[HTML]{4472C4} \textbf{0.60}} & {\color[HTML]{4472C4} 1.58} & {\color[HTML]{4472C4} 0.92} & {\color[HTML]{4472C4} 1.09} & {\color[HTML]{4472C4} 1.17} & {\color[HTML]{4472C4} 1.07} & {\color[HTML]{4472C4} \textbf{0.55}} & {\color[HTML]{4472C4} \textbf{0.62}} & {\color[HTML]{4472C4} \textbf{0.36}} & {\color[HTML]{4472C4} \textbf{0.71}} & {\color[HTML]{4472C4} \textbf{0.79}} & {\color[HTML]{4472C4} \textbf{1.51}} & {\color[HTML]{4472C4} \textbf{0.75}} & {\color[HTML]{4472C4} \textbf{1.27}} & {\color[HTML]{4472C4} 0.67} & {\color[HTML]{4472C4} 2.05} & {\color[HTML]{4472C4} \textbf{0.89}} & {\color[HTML]{4472C4} \textbf{1.386}} & {\color[HTML]{4472C4} \textbf{0.775}} \\ \hline
\end{tabular}} 
\end{center}
\end{table*}

\begin{table*}[!t]
\setlength{\belowcaptionskip}{0cm}
\setlength{\abovecaptionskip}{-0.5cm}
\caption{Ford odometry evaluation.}\label{tab2}
\begin{center}
\resizebox{1.0\textwidth}{!}{  
\begin{tabular}{c|cc|cc|cc|cc|cc||cc|cc}
\hline
                       & \multicolumn{2}{c|}{ICP-po2po} & \multicolumn{2}{c|}{ICP-po2pl} & \multicolumn{2}{c|}{GICP \cite{GICP}} & \multicolumn{2}{c|}{CLS \cite{CLS}} & \multicolumn{2}{r||}{LOAM(w/mapping) \cite{Loam}} & \multicolumn{2}{c|}{LO-Net \cite{Lo-net}}  & \multicolumn{2}{c}{{\color[HTML]{4472C4} Ours}}                     \\ \cline{2-15} 
\multirow{-2}{*}{Seq.} & $t_{rel}$      & $r_{rel}$     & $t_{rel}$      & $r_{rel}$     & $t_{rel}$   & $r_{rel}$   & $t_{rel}$   & $r_{rel}$  & $t_{rel}$         & $r_{rel}$        & $t_{rel}$     & $r_{rel}$     & {\color[HTML]{4472C4} $t_{rel}$} & {\color[HTML]{4472C4} $r_{rel}$} \\ \hline
Ford-1                 & 8.2            & 2.64          & 3.35           & 1.65          & 3.07        & 1.17        & 10.54       & 3.9        & \textbf{1.68}     & \textbf{0.54}    & \textbf{2.27} & \textbf{0.62} & {\color[HTML]{4472C4} 2.99}      & {\color[HTML]{4472C4} 1.83}      \\
Ford-2                 & 16.23          & 2.84          & 5.68           & 1.96          & 5.11        & 1.47        & 14.78       & 4.6        & \textbf{1.78}     & \textbf{0.49}    & \textbf{2.18} & \textbf{0.59} & {\color[HTML]{4472C4} 3.75}      & {\color[HTML]{4472C4} 1.38}      \\ \hline
\end{tabular}} 
\end{center}
\end{table*}

\begin{table*}[!t]
\setlength{\belowcaptionskip}{0cm}
\setlength{\abovecaptionskip}{-0.5cm}
\caption{Apollo odometry evaluation.}\label{tab3}
\begin{center}
\resizebox{1.0\textwidth}{!}{
\begin{tabular}{c|cc|cc|cc|cc|cc||cc|cc}
\hline
                      & \multicolumn{2}{c|}{ICP-po2po} & \multicolumn{2}{c|}{ICP-po2pl} & \multicolumn{2}{c|}{GICP \cite{GICP}} & \multicolumn{2}{c|}{NDT-P2D \cite{NDT}} & \multicolumn{2}{c||}{LOAM(w/mapping) \cite{Loam}} & \multicolumn{2}{c|}{SelfVoxeLO \cite{SelfVoxeLO}} & \multicolumn{2}{c}{{\color[HTML]{4472C4} Ours}}                             \\ \cline{2-15} 
                      & $t_{rel}$      & $r_{rel}$     & $t_{rel}$      & $r_{rel}$     & $t_{rel}$   & $r_{rel}$   & $t_{rel}$     & $r_{rel}$    & $t_{rel}$         & $r_{rel}$        & $t_{rel}$       & $r_{rel}$      & {\color[HTML]{4472C4} $t_{rel}$}     & {\color[HTML]{4472C4} $r_{rel}$}     \\ \hline
Test mean             & 22.8           & 2.35          & 7.75           & 1.2           & 4.55        & 0.76        & 57.2          & 9.4          & \textbf{5.93}     & \textbf{0.26}    & 6.42            & 1.65           & {\color[HTML]{4472C4} \textbf{4.53}} & {\color[HTML]{4472C4} \textbf{1.30}} \\ \hline
\end{tabular}} 
\end{center}
\end{table*}
\subsection{Loss Function}
For unsupervised training, We use the same point cloud down-sampling strategy 
as \cite{My}. we fisrt use RANSAC \cite{RANSAM} to remove the ground points more 
accurately. Next we also perform voxel downsample. The form of loss is 
similar to \cite{Cho}, but we only use point-to-plane ICPloss. 
For the source point cloud $\overline{S}_k$, find the closest 
point in the target point cloud $\overline{S}_{k+1}$ as the matching point. 
The formula of loss function $L$ is shown as 
\begin{equation}
\begin{split}
L=&\{\sum_{i=0}^3\sum_{\overline{s}_k\in\overline{S}_k}
(R_i\cdot \overline{s}_k+T_i-\overline{s}_{k+1})\cdot\overline{n}_{k+1}|\\
&\mathop{\min}\limits_{\overline{s}_{k+1}}
(R_i\cdot \overline{s}_k+T_i-\overline{s}_{k+1})^2,\\
&\overline{s}_{k+1}\in\overline{S}_{k+1},
\overline{n}_{k+1}\in\overline{N}_{k+1}\},
\end{split}
\end{equation}
where $\overline{N}_{k+1}$ is the normal vector of the target point cloud.  

\section{EXPERIMENTAL EVALUATION}
In this section, we first introduce the implementation details of our 
total framework and three public datasets we use in experiment. Then, we 
compare our method with other existing lidar odometry methods to show 
that our model is competitive. Finally, we perform ablation studies to 
verify the effectiveness of the innovative part of our model.
\subsection{Implementation Details}
Our proposed model is implemented in PyTorch \cite{Pytorch}. We train 
it with a single NVIDIA Titan RTX GPU. We optimize the parameters with the 
Adam optimizer \cite{Adam} with hyperparameter values of $\beta_1 = 0.9$, 
$\beta_2 = 0.99$ and $w_{decay} = 10^{-5}$. We adopt step scheduler with 
different stepsizes (20 on KITTI \cite{KITTI} and Ford \cite{Ford} , 10 
on Apollo \cite{Apollo}) and $\gamma = 0.5$ to control the training procedure 
with initial learning rate of $10^{-3}$. The batch size is also different 
in different datasets(10 on KITTI \cite{KITTI} and Ford \cite{Ford}, 
16 on Apollo \cite{Apollo}). During data preprocessing, we drop $U=50\%$ 
points and side length of voxel cell $D=0.3m$. In order to simplify the 
training process, we unify the number of voxel cell to $C^0=8192$. If the number 
of points is too small, the existing points will be copied  
to fill them. If there are too many points, the voxel cells with fewer 
points will be removed prior. We reserve $C^1=2048,C^2=512,C^3=256$ points 
for other level of point cloud.

\begin{figure}[!t]
        \includegraphics[width=\linewidth]{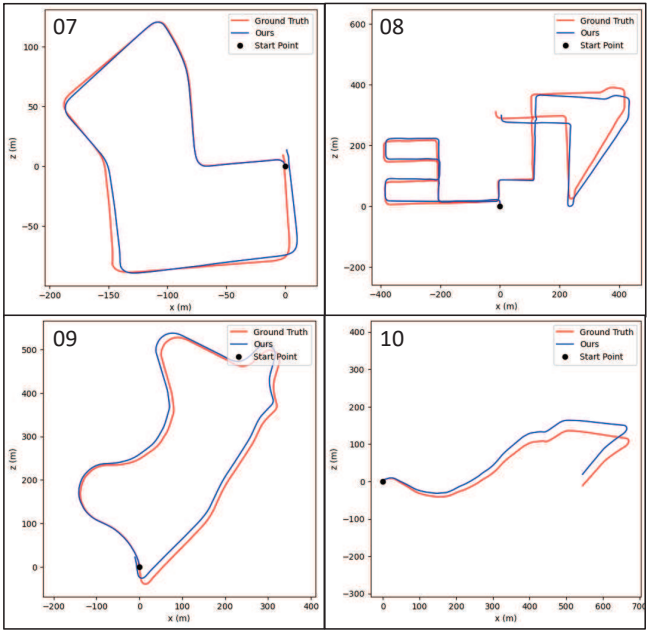}
        \caption{2D estimated trajectories of our method on sequence 07-10.} 
        \label{fig5}
\end{figure}

\subsection{Datasets} 
\subsubsection{KITTI}
The KITTI odometry dataset \cite{KITTI} is the most classic data set that 
most odometry methods will test on it. It has 22 different sequences, only 
Sequences 00-10 have an official public ground truth. We use sequences 00-06 
for training and 07-10 for testing which is the splitting strategy with 
minimal training data in all other deep learning method.
\subsubsection{Ford Campus Vision and Lidar DataSet}
Ford \cite{Ford} only contains two sequences, but there are more moving 
vehicles than the KITTI dataset which make it more difficult. In order to 
compare with Lo-net \cite{Lo-net}, we do not train on it, but use the model 
trained on KITTI to test directly. 
\subsubsection{Apollo-SouthBay Dataset}
Apollo-SouthBay Dataset \cite{Apollo} collected in the San Francisco 
Bay area,United States, covers various scenarios. It provides ground truth 
in all sequences and splits them into traindata, testdata and mapdata three 
parts. In our experiment, traindata and testdata set are respectively used 
for train and test which is the same with SelfVoxeLO \cite{SelfVoxeLO}. 
It is more complicated and longer than KITTI.

\begin{table*}[!t]
\setlength{\belowcaptionskip}{0cm}
\setlength{\abovecaptionskip}{-0.5cm}
\caption{The ablation study results on KITTI.}\label{tab4}
\begin{center}
\resizebox{1.0\textwidth}{!}{
\begin{tabular}{c|cc|cc|cc|cc|cc|cc|cc|cc|cc|cc|cc|cc}
\hline
                                               & \multicolumn{2}{c|}{00*}                                                    & \multicolumn{2}{c|}{01*}                                  & \multicolumn{2}{c|}{02*}                                                    & \multicolumn{2}{c|}{03*}                                                    & \multicolumn{2}{c|}{04*}                                           & \multicolumn{2}{c|}{05*}                                           & \multicolumn{2}{c|}{06*}                                                    & \multicolumn{2}{c|}{07}                                            & \multicolumn{2}{c|}{08}                                                     & \multicolumn{2}{c|}{09}                                                     & \multicolumn{2}{c|}{10}                                                     & \multicolumn{2}{c}{Mean on 07-10}                                                   \\ \cline{2-25} 
\multirow{-2}{*}{Method}                       & $t_{rel}$                            & $r_{rel}$                            & $t_{rel}$                   & $r_{rel}$                   & $t_{rel}$                            & $r_{rel}$                            & $t_{rel}$                            & $r_{rel}$                            & $t_{rel}$                            & $r_{rel}$                   & $t_{rel}$                            & $r_{rel}$                   & $t_{rel}$                            & $r_{rel}$                            & $t_{rel}$                   & $r_{rel}$                            & $t_{rel}$                            & $r_{rel}$                            & $t_{rel}$                            & $r_{rel}$                            & $t_{rel}$                            & $r_{rel}$                            & $t_{rel}$                             & $r_{rel}$                             \\ \hline
MLP                                            & 3.67                                 & 1.67                                 & 4.08                        & 1.11                        & 4.21                                 & 1.75                                 & 3.08                                 & 1.26                                 & 2.09                                 & 1.48                        & 2.39                                 & 1.28                        & 1.57                                 & 0.82                                 & 17.02                       & 9.38                                 & 9.82                                 & 4.61                                 & 9.82                                 & 4.61                                 & 22.06                                & 11.53                                & 14.677                                & 7.530                                 \\
SVD-po2po                                      & 1.73                                 & 0.80                                 & 3.70                        & 1.05                        & 2.25                                 & 0.89                                 & 2.14                                 & 1.22                                 & 1.02                                 & \textbf{0.50}               & 1.24                                 & 0.70                        & 0.81                                 & 0.49                                 & 6.22                        & 4.62                                 & 5.11                                 & 2.14                                 & 9.77                                 & 3.57                                 & 8.54                                 & 4.04                                 & 7.410                                 & 3.594                                 \\
SVD-po2pl                                      & 1.51                                 & 0.78                                 & \textbf{3.51}               & \textbf{0.92}               & 2.06                                 & 0.87                                 & 2.28                                 & 1.12                                 & 1.52                                 & 1.17                        & 1.09                                 & 0.68                        & 0.90                                 & 0.63                                 & 3.34                        & 1.81                                 & 3.14                                 & 1.21                                 & 3.48                                 & 1.53                                 & 4.59                                 & 1.92                                 & 3.639                                 & 1.619                                 \\
Muti-SVD-po2pl w/o refine                      & 1.47                                 & 0.77                                 & 3.54                        & 0.97                        & 1.75                                 & 0.71                                 & 2.34                                 & 1.11                                 & 1.08                                 & 0.92                        & 0.98                                 & \textbf{0.52}               & 0.73                                 & 0.37                                 & \textbf{1.23}               & 1.02                                 & 1.91                                 & 0.85                                 & 2.26                                 & 0.96                                 & 2.25                                 & 1.13                                 & 1.911                                 & 0.990                                 \\
{\color[HTML]{4472C4} Muti-SVD-po2pl w refine} & {\color[HTML]{4472C4} \textbf{1.35}} & {\color[HTML]{4472C4} \textbf{0.68}} & {\color[HTML]{4472C4} 3.53} & {\color[HTML]{4472C4} 1.00} & {\color[HTML]{4472C4} \textbf{1.56}} & {\color[HTML]{4472C4} \textbf{0.63}} & {\color[HTML]{4472C4} \textbf{1.47}} & {\color[HTML]{4472C4} \textbf{0.80}} & {\color[HTML]{4472C4} \textbf{0.96}} & {\color[HTML]{4472C4} 0.90} & {\color[HTML]{4472C4} \textbf{1.03}} & {\color[HTML]{4472C4} 0.53} & {\color[HTML]{4472C4} \textbf{0.67}} & {\color[HTML]{4472C4} \textbf{0.36}} & {\color[HTML]{4472C4} 1.40} & {\color[HTML]{4472C4} \textbf{0.77}} & {\color[HTML]{4472C4} \textbf{1.62}} & {\color[HTML]{4472C4} \textbf{0.64}} & {\color[HTML]{4472C4} \textbf{2.00}} & {\color[HTML]{4472C4} \textbf{0.84}} & {\color[HTML]{4472C4} \textbf{1.49}} & {\color[HTML]{4472C4} \textbf{1.11}} & {\color[HTML]{4472C4} \textbf{1.626}} & {\color[HTML]{4472C4} \textbf{0.840}} \\ \hline
\end{tabular}} 
\end{center}
\end{table*}

\subsection{Evaluation Result}
We adopt the official evaluation metrics provide by the KITTI benchmark 
\cite{KITTI} on all our experiments (include other two datasets). It contains 
two key indicators: $t_{rel}$ means average translational RMSE (\%) on length 
of 100m-800m and $r_{rel}$ means average rotational RMSE ($^\circ$/100m) on 
length of 100m-800m. The smaller they are, the better the method is. \\
\indent We compare our method with the following methods which can be 
divided into two types. Model-based methods are: ICP-po2po, ICP-po2pl, 
GICP \cite{GICP}, CLS \cite{CLS} and LOAM \cite{Loam}. Learning-based methods 
are: Lo-Net \cite{Lo-net}, SelfVoxeLO \cite{SelfVoxeLO}, Nubert \emph{et al.} 
\cite{Nubert} and Cho \emph{et al.} \cite{Cho}. only Lo-Net is supervised 
method, others are unsupervised/self-supervised methods.

\subsubsection{Evaluation on KITTI}
Learning-based methods adopted different splitting strategies for training and 
testing on KITTI. Nubert \emph{et al.} and Cho \emph{et al.} \cite{Cho} train
on sequences 00-08 and test sequences on 09-10. Others train
on sequences 00-06 and test sequences on 07-10. Some of they do not show all
rsults in their paper and do not release their code, so we use "NA" represent
result we unknow. \\ 
\indent Quantitative results are listed in Table.~\ref{tab1}. As is shown in table, 
compared with learing-based methods, the results of our method are far better 
than all current unsupervised methods, even if compared with supervised methods 
Lo-net \cite{Lo-net}, we can also even be comparable. Compared with 
model-based method, we only lose to GISP which with slow calculation speed 
and the most classic lidar SLAM algorithm LOAM which with the additional mapping 
step. Fig.~\ref{fig5} plots trajectories in sequences 07-10 of our methods for
 visualization. 

\subsubsection{Evaluation on Ford}
Quantitative results are listed in Table.~\ref{tab2}. Both Lo-net \cite{Lo-net}
and us only train on KITTI and test on Ford. Due to Lo-net \cite{Lo-net} is a
supervised method, our results are not as good as him, but the gap is not too 
big. Compared with model-based method, our method only loses to LOAM \cite{Loam}.
In addition, our method surpasses GICP \cite{GICP} slightly that we inferior 
to it in KITTI.

\subsubsection{Evaluation on Apollo}
Apollo-SouthBay dataset \cite{Apollo} is longer and hard than above two dataset, 
so it can further demonstrate generality of our method. Quantitative results also are 
listed in Table.~\ref{tab3}. Not surprisingly, our method can also greatly surpasses 
the state-of-the-art method  with the same training set and test set. Compared with 
the model-based method, for the first time, we surpassed all methods including LOAM 
\cite{Loam} in $t_{rel}$, and lower than LOAM \cite{Loam}and GICP \cite{GICP} 
on $t_{rel}$.

\begin{figure}[!t]
        \includegraphics[width=\linewidth]{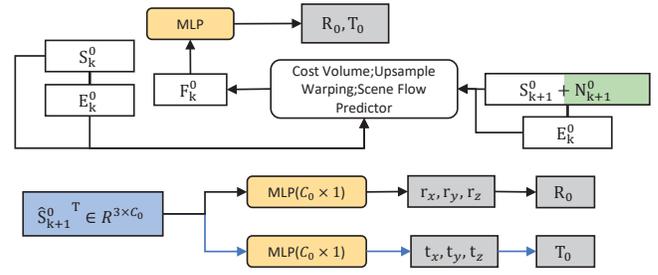}
        \caption{"MLP" detail structure.} 
        \label{fig6}
\end{figure}
\begin{figure}[!t]
        \includegraphics[width=\linewidth]{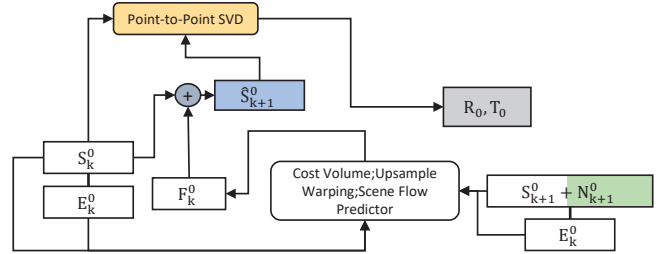}
        \caption{"SVD po2po" detail structure.} 
        \label{fig7}
\end{figure}
\begin{figure}[!t]
        \includegraphics[width=\linewidth]{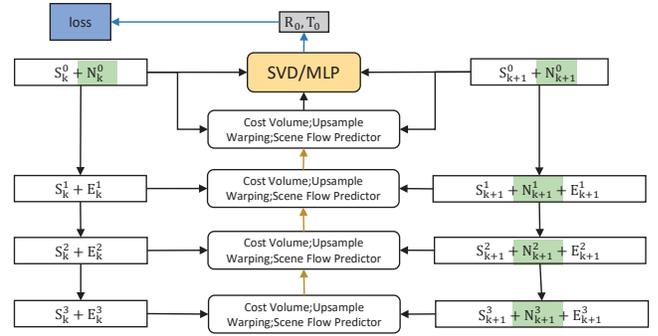}
        \caption{Only use SVD once in Level-0.} 
        \label{fig8}
\end{figure}
\begin{figure}[!t]
        \includegraphics[width=\linewidth]{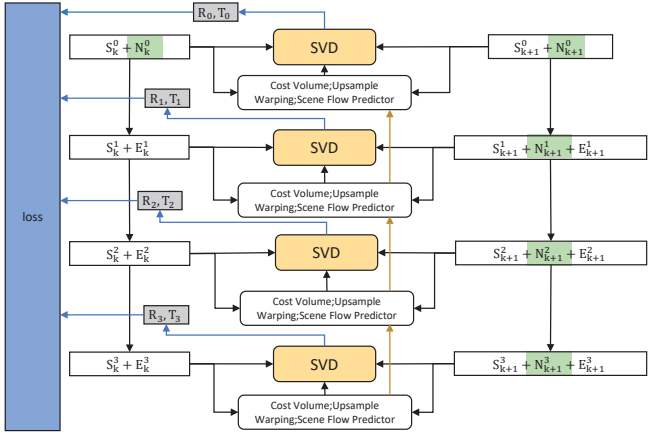}
        \caption{Not refine point clouds.} 
        \label{fig9}
\end{figure}
\subsection{Ablation Study}
In order to prove the effectiveness of our improvement based on the PointPWC 
\cite{PointPWC}, as shown in the Table.~\ref{tab4}, we have taken a series of 
ablation experiments on the KITTI \cite{KITTI}. We test the effect of using 
different methods to transform the scene flow into pose, and use the simplest 
method as directly inputing the scene flow into the fully connected layer and 
output $r^{x,y,z}$ and $t^{x,y,z}$ as benchmarks whose method
name is "MLP" (see Figure.~\ref{fig6}). We adopt SVD on point-to-point ICP 
method \cite{SVD} which named as "SVD-po2po" (see Figure.~\ref{fig7}), the 
lidar odometry method DMLO \cite{DMLO} and the point cloud registration method 
PointDSC \cite{PointDSC} use in thier method. Next We adopt SVD on 
point-to-plane \cite{po2plICP} which has been used in our method and we name 
it as "SVD-po2pl" (similar to Figure.~\ref{fig4}, only the level is different, 
it is 0 in this part not 3). 
In the previous experiments, pose is calculated only after 
the scene flow is obtained at the lowest level, and is not used in the high level. 
(see Figure.~\ref{fig8}). In method "Muti-SVD-po2pl w/o refine", SVD is performed 
in each layer to solve the pose, but the solved pose is not used to improve the 
low-level point cloud (see Figure.~\ref{fig9}). Finally, the method "Muti-SVD-po2pl 
w refine" we propose not only obtains the pose of each layer, but also  
refines low level point cloud. It can be seen that every improvement in our
network makes the test results better, which proves that improvement is effective. 
\section{CONCLUSIONS}
We proposed a new lidar odometry framework based on the scene flow estimate 
network PointPWC \cite{PointPWC}. Different from other lidar odometry methods, 
we not only extracted point-level features instead of global-level features, 
but also use SVD decomposition to convert the scene flow into pose, and additionally use 
normal vector to convert point-to-point ICP into point-to-plane ICP which further 
improves the accuracy. We also use high-level pose to improve the low-level 
point cloud, so that the pose of each layer is associated with each other. 
The idea of iterative optimization of the pose in the traditional algorithm is 
integrated into our network. In the future, we will explore the combination of 
our method and mapping optimization to achieve further breakthroughs. 
\addtolength{\textheight}{-12cm}   % This command serves to balance the column lengths

%%%%%%%%%%%%%%%%%%%%%%%%%%%%%%%%%%%%%%%%%%%%%%%%%%%%%%%%%%%%%%%%%%%%%%%%%%%%%%%%

%%%%%%%%%%%%%%%%%%%%%%%%%%%%%%%%%%%%%%%%%%%%%%%%%%%%%%%%%%%%%%%%%%%%%%%%%%%%%%%%

%%%%%%%%%%%%%%%%%%%%%%%%%%%%%%%%%%%%%%%%%%%%%%%%%%%%%%%%%%%%%%%%%%%%%%%%%%%%%%%%

%%%%%%%%%%%%%%%%%%%%%%%%%%%%%%%%%%%%%%%%%%%%%%%%%%%%%%%%%%%%%%%%%%%%%%%%%%%%%%%%
\bibliography{reference}
\bibliographystyle{IEEEtranS}
\end{document}